\documentclass[10pt,twocolumn,letterpaper]{article}

\usepackage[pagenumbers]{iccv} 

\usepackage{bm}

\usepackage{subcaption}
\usepackage{pifont}

\usepackage{multirow}
\usepackage{booktabs}
\usepackage{amsmath}
\usepackage{colortbl}
\usepackage{tcolorbox}

\definecolor{iccvblue}{rgb}{0.21,0.49,0.74}
\usepackage[pagebackref,breaklinks,colorlinks,allcolors=iccvblue]{hyperref}

\def\paperID{13198} 
\def\confName{ICCV}
\def\confYear{2025}
\def\titleDef{\textbf{\textit{InstaDrive}}}

\title{InstaDrive: Instance-Aware Driving World Models for Realistic and Consistent Video Generation}

\author{
Zhuoran Yang$^{1}$\hspace{0.8em}
Xi Guo$^{2}$\hspace{0.8em}
Chenjing Ding$^{2}$ \hspace{1em}
Chiyu Wang$^2$\hspace{0.8em}
Wei Wu$^{2,3}$
Yanyong Zhang$^{1}$\thanks{Corresponding Author}
\\
{
 \normalsize  $^1$University of Science and Technology of China}  \enspace
 \normalsize $^2$SenseAuto \enspace \normalsize $^3$Tsinghua University \\
{{\tt\small shanpoyang@mail.ustc.edu.cn}, \enspace \tt\small \{guoxi,dingchenjing\}@sensetime.com}  \\
{\tt\small \{wangchiyu, wuwei\}@senseauto.com} \\
{\tt\small \{yanyongz\}@ustc.edu.cn} \\
}

\begin{document}
\maketitle
\begin{abstract}
Autonomous driving relies on robust models trained on high-quality, large-scale multi-view driving videos. 
While world models offer a cost-effective solution for generating realistic driving videos, they struggle to maintain instance-level temporal consistency and spatial geometric fidelity.
To address these challenges, we propose \titleDef, a novel framework that enhances driving video realism through two key advancements: (1) Instance Flow Guider, which extracts and propagates instance features across frames to enforce temporal consistency, preserving instance identity over time.
(2) Spatial Geometric Aligner, which improves spatial reasoning, ensures precise instance positioning, and explicitly models occlusion hierarchies.
By incorporating these instance-aware mechanisms, \titleDef~ achieves state-of-the-art video generation quality and enhances downstream autonomous driving tasks on the nuScenes dataset.
Additionally, we utilize CARLA's autopilot to procedurally and stochastically simulate rare but safety-critical driving scenarios across diverse maps and regions, enabling rigorous safety evaluation for autonomous systems.
Our project page\footnote{\url{https://shanpoyang654.github.io/InstaDrive/page.html}}.
\end{abstract}    
\section{Introduction}
\label{section:intro}

\begin{figure*}[t]
\centering
\includegraphics[width=0.75\linewidth]{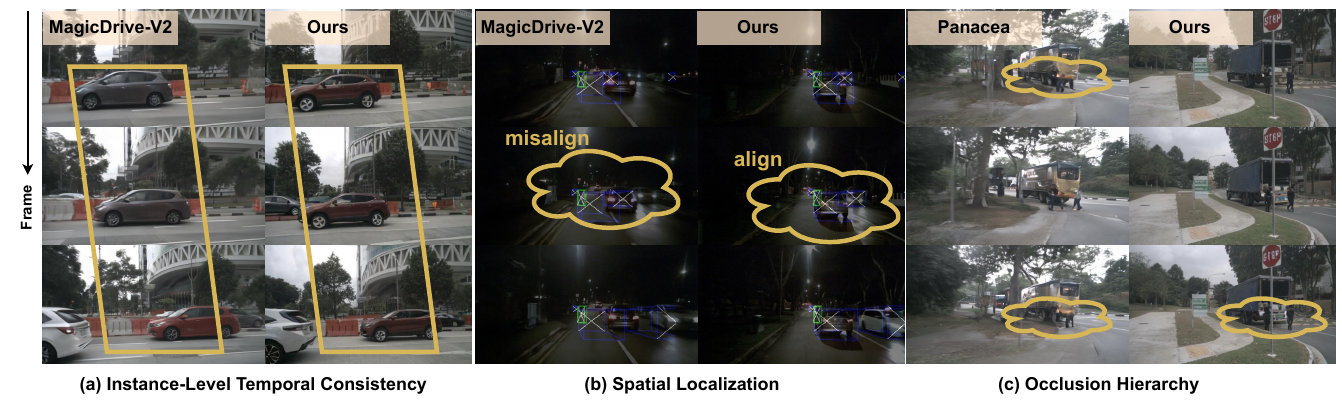} 
\caption{
\textbf{Limitations of Prior Works.}
\textbf{(a)} Temporal Consistency: In MagicDrive-V2 \cite{gao2024magicdrivedithighresolutionlongvideo}, the car’s color changes inconsistently over time.
\textbf{(b)} Spatial Localization: In MagicDrive-V2, the car deviates from the bounding box control signal. 
\textbf{(c)} Occlusion Hierarchy: In Panacea \cite{wen2024panacea}, the distant bus incorrectly occludes the nearby pedestrian, violating natural occlusion rules.
Our method excels in temporal consistency, spatial localization, and occlusion hierarchy, addressing these issues effectively.
 }
\label{fig:intro}
\end{figure*}

Autonomous driving has attracted extensive attention from both industry and academia for decades \cite{shi2016uniad, zheng2024genad, chen2024end, jiang2023vad}. 
To enhance the performance and reliability of autonomous systems, high-quality, large-scale multi-view driving videos with precise annotations are essential for training models on downstream tasks such as perception, object tracking, and planning. 
However, acquiring and labeling real-world driving data is both costly and labor-intensive. 
To address this, benefiting from the rapid advancements in video generation models \cite{lei2023pyramidflow, xi2025sparse, zheng2024open, hacohen2024ltx, wang2024qihoo, gao2024ca2, zhou2024allegro, ho2022video, blattmann2023align, hu2024animate, wang2023modelscope, wang2024lavie, bar2024lumiere, gupta2024photorealistic}, driving world models \cite{drivedreamer2, wen2024panacea, jia2023adriver, wang2023drive, gao2024magicdrivedithighresolutionlongvideo} have emerged as a promising solution, capable of generating diverse and realistic driving scenarios while significantly reducing data collection and annotation costs.

Instance-level temporal consistency and spatial geometric fidelity are critical for generating realistic driving videos, as they affect video quality \cite{Thomas2018fvd} and directly impact their applicability in autonomous driving tasks. Multi-object Tracking \cite{streampetr} and planning \cite{hu2023planningorientedautonomousdriving} require generated driving videos with temporally stable instance appearances to enhance temporal context understanding. This necessitates that the world model accurately maintains instance identities across frames, ensuring continuity in the motion and behavior of surrounding objects.
On the other hand, perception tasks \cite{streampetr} require generated driving videos where instance positioning strictly adheres to spatial constraints imposed by bounding box control signals, ensuring accurate spatial context comprehension. This requires the world model to accurately capture instance spatial locations and occlusion hierarchy, ensuring geometric consistency within the scene.
These two factors are essential for enabling world models to effectively learn the underlying dynamics of real-world environments. From a technical standpoint, ensuring strong temporal consistency and precise spatial alignment enhances the reliability of autonomous driving models trained on synthetic data, ultimately improving their real-world performance.

However, generating driving videos that maintain instance-level temporal consistency and spatial geometric fidelity remains two significant challenges, primarily due to the large sampling space in diffusion-based models and the limited control signals. 
First, maintaining fine-grained temporal consistency is particularly difficult. While prior works \cite{gao2023magicdrive, gao2024magicdrivedithighresolutionlongvideo, wen2024panacea, drivedreamer2, wang2023drive, li2023drivingdiffusionlayoutguidedmultiviewdriving} have incorporated various techniques to improve \textit{global} coherence, they still struggle with \textit{instance-level} temporal inconsistencies. For instance, DriveDreamer \cite{wang2023drive}, MagicDrive-V2 \cite{gao2024magicdrivedithighresolutionlongvideo}, and Panacea \cite{wen2024panacea} integrate temporal attention layers to enhance inter-frame consistency.
However, without explicit control mechanisms to enforce instance-level consistency, these methods struggle to maintain stable instance attributes across frames, often resulting in color shifts or texture inconsistencies in Fig.\ref{fig:intro} (a). This highlights the need for explicit instance-level control signals to improve temporal consistency.
Second, achieving spatial geometric fidelity presents another challenge. 
Existing methods often suffer from spatial misalignment of instances, as seen in MagicDrive-V2 \cite{gao2024magicdrivedithighresolutionlongvideo}, where instance locations exhibit noticeable deviations in Fig.\ref{fig:intro} (b). This issue arises because these methods lack explicit view transformation from BEV to the camera’s First-Person View (FPV). 
Furthermore, existing methods struggle to capture the occlusion hierarchy among instances in Fig.~\ref{fig:intro} (c), where the lack of explicit depth order prevents correct occlusion reasoning. This highlights the need for an improved spatial control mechanism that ensures accurate instance positioning and explicitly models the occlusion hierarchy.

In this paper, to address the above challenges, we propose \titleDef, a driving world model that effectively adheres to instance-level temporal consistency and spatial geometric fidelity. \titleDef~ achieves state-of-the-art performance in both video quality and validation in downstream autonomous driving tasks.
To ensure instance-level temporal consistency, we introduce \textbf{I}nstance \textbf{F}low \textbf{G}uider (\textbf{IFG}), a lightweight module to extract and propagate instance features across frames. We propose a simple but effective instance flow for mapping the object motion to the RGB domain, then inject the flow to DiT backbone. IFG provides an instance-aware motion cue, which serves as a reference for tracking the position of the corresponding instances in the previous frame. Once the position is determined, we retrieve the semantic features of the instance, such as object category and color. These features are then propagated forward to ensure instance-level visual consistency across frames. This mechanism significantly improves instance-level temporal consistency.

To ensure instance-level spatial geometric fidelity, we introduce \textbf{S}patial \textbf{G}eometric \textbf{A}ligner (\textbf{SGA}), which enables the model to accurately capture the spatial locations of the instance and the occlusion hierarchy. 
To enable accurate spatial localization, we transform 3D bounding boxes into the camera's perspective view, extracting projected instance bounding boxes as control elements. This transformation leverages the camera's intrinsic and extrinsic parameters, ensuring spatial consistency between the world coordinate system and the image plane.
To establish a consistent occlusion hierarchy, we explicitly resolve the relative depth order of instances using the distance from each corner point to the camera's optical center, measured along the optical axis. We encode each corner point's location and depth via Fourier embedding, followed by an MLP, to derive an explicit depth order representation.
Through this mechanism, the model effectively learns to capture spatial locations and occlusion relationships between instances.

\titleDef~ establishes a fine-grained and robust world model which ensures instance-level temporal consistency and geometric fidelity in generated driving videos. 
Our approach achieves state-of-the-art performance in both video generation quality and downstream autonomous driving task validation, outperforming previous works \cite{gao2024magicdrivedithighresolutionlongvideo, drivedreamer2, li2023drivingdiffusionlayoutguidedmultiviewdriving, wen2024panacea}. 
Our contributions are as follows. 
\begin{itemize}    
    \item To maintain stable instance attributes over time, we propose the Instance Flow Guider, which extracts and propagates instance features across frames, enabling instance-level temporal consistency.

    \item To enable accurate spatial location and establish a consistent occlusion hierarchy, we introduce the Spatial Geometric Aligner, which integrates the 3D bounding boxes and their occlusion relationships, and aligns these details with the pixel space.
    
    
    \item Our model achieves SOTA video generation quality with high FID and FVD on the nuScenes benchmark, surpassing previous methods. For autonomous driving applications, the generated videos are validated on downstream perception, tracking, and planning tasks, with performance competitive to real-world sensor data.

    \item We designed a pipeline that generates virtual layouts based on CARLA and employs our model as a renderer for video synthesis, demonstrating the immense potential of \titleDef~ in generating long-tail driving scenarios.

\end{itemize}

\section{Related Works}
\label{sec:related}

\noindent \textbf{Controllable Generation.}  
The emergence of diffusion models \cite{zhang2024moonshot} has significantly advanced text-to-video generation \cite{an2023latent, blattmann2023align, guo2023animatediff, he2022latent, ho2022imagen, svd, singer2022make, wang2023videofactory, zhou2023magicvideo, cogvideox, guo2024infinitydrive}. Video LDM \cite{blattmann2023align} accelerates generation by denoising in the latent space, but text prompts alone lack precision. Recent methods combine image blocks with text for better control \cite{zhang2024moonshot}. Our work focuses on generating realistic street-view videos, addressing challenges like complex layouts and dynamic vehicles. We enhance control by integrating road maps, 3D bounding boxes, and BEV keyframes.

\noindent \textbf{Street-View Generation.}  
Street-view generation methods typically use 2D layouts like BEV maps, 2D bounding boxes, and semantic segmentation. BEVGen \cite{swerdlow2023street} encodes semantic data in BEV layouts, while BEVControl \cite{bevcontrol} uses a two-stage pipeline for multi-view urban scenes, ensuring cross-view consistency. However, projecting 3D information into 2D layouts loses geometric details, causing temporal inconsistencies in videos. To address this, we use 3D bounding boxes to maintain geometric fidelity. Unlike DrivingDiffusion \cite{li2023drivingdiffusionlayoutguidedmultiviewdriving}, which relies on a complex multi-stage pipeline, our method simplifies the process with an efficient, end-to-end framework, ensuring temporal coherence and computational efficiency.

\noindent \textbf{Multi-View Video Generation.}  
Multi-view video generation faces challenges in achieving both multi-view and temporal consistency. MVDiffusion \cite{tang2023mvdiffusion} uses a correspondence-aware attention module to align views, while Tseng et al. \cite{poseguideddiffusion} apply epipolar geometry for view-to-view regularization. DriveScape \cite{drivescape}, MagicDrive \cite{gao2023magicdrive}, and MagicDrive V2 \cite{gao2024magicdrivedithighresolutionlongvideo} incorporate advanced control signals, high-definition rendering, and long-duration training but struggle with instance-level temporal consistency and precise positional control.

\noindent \textbf{Simulation-to-Real Visual Translation.}  
Recent advances in synthetic data for real-world visual tasks have shown significant progress. GAN-based translation \cite{guo2020gan} and domain randomization \cite{tobin2017domain} bridge synthetic and real-world data distributions, while datasets like Synthia \cite{ros2016synthia} and Virtual KITTI \cite{cabon2020virtual} provide scalable benchmarks for semantic segmentation and autonomous driving. Adversarial training \cite{shrivastava2017learning, zhang2018collaborative} reduces distribution gaps, and human motion representation learning \cite{guo2022learning} highlights synthetic data’s utility in video understanding. Unlike these methods, we extract proxy data like 3D bounding boxes and road maps from graphics systems, leveraging the InstaDrive model to generate more realistic and diverse videos.

\section{Method}
\label{section:method}
\begin{figure*}[ht]
\centering
\includegraphics[width=0.85\linewidth]{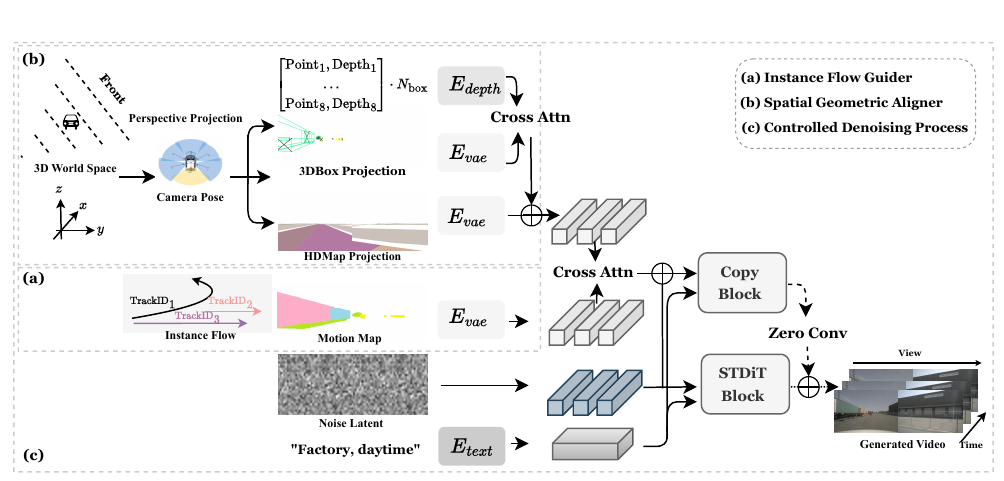} 
\caption{\textbf{Overview.}
(a) Instance Flow Guider, which utilizes the instance flow to improve instance-level temporal consistency.
(b) Spatial Geometric Aligner, which uses perspective projection and depth order to capture instance spatial locations and occlusion hierarchy. 
(c) Controlled Denoising Process, enabled by ST-DiT with ControlNet for unified condition control.}
\label{fig: main}
\vspace{-0.5cm}
\end{figure*}

We introduce \titleDef~ in Sec. \ref{section:InstaDrive}, a novel framework for generating realistic driving videos that adhere to instance-level temporal consistency and spatial geometric fidelity. 
Furthermore, we leverage CARLA's autopilot to procedurally and stochastically simulate rare yet safety-critical driving scenarios across diverse maps and regions in Sec. \ref{section:carla}.

\subsection{Overview} 
\label{section:InstaDrive}
The overall architecture of our model is illustrated in Fig. \ref{fig: main}.
Building on OpenSora V1.1 ~\cite{opensora}, we employ a Variational Auto-Encoder (VAE) for video encoding, T5 ~\cite{raffel2020exploring} for text encoding, and Spatial-Temporal Diffusion Transformer (ST-DiT) as foundational model for the denoising process.

To achieve fine-grained control over both foreground and background elements, we introduce a comprehensive set of control conditions, including bounding box projection, road maps, camera poses, and scene descriptions, integrating them into the conditioned video generation process.
Moreover, we introduce instance tracking IDs to facilitate the tracking and propagation of instance features across frames, a key mechanism within our proposed Instance Flow Guider module, detailed in Sec. \ref{section: dynamic}.
Additionally, we leverage box coordinate and bounding box projection to enforce instance-level spatial geometric fidelity, as elaborated in Sec. \ref{section: static}.

Given the need for handling multiple control elements, we employ ControlNet~\cite{zhang2023adding} to inject control signals into the video generation process.
Practically, a set of encoders, including $E_{depth}$, $E_{vae}$, and $E_{text}$, extract latent features from diverse control conditions.
To incorporate these control-aware representations, we integrate 13 duplicated blocks into the first 13 base blocks of the ST-DiT architecture.
Each control block modulates the feature flow by fusing condition features with the corresponding base block outputs, thereby ensuring effective control signal conditioning throughout the generation pipeline.
 
To guarantee the multi-view consistency during generation, we use a parameter-free view-inflated attention mechanism to replace the commonly used cross-view attention modules~\cite{gao2023magicdrive}.
Specifically, we reshape the input from $\mathbb{R}^{v \times t \times h \times w \times c}$ to $\mathbb{R}^{t \times h \times (wv) \times c}$ and treat $wv$ as the frame width.
Our proposed approach improves the multi-view coherence without compensating with additional parameters.

\begin{figure}[t]
\centering
\includegraphics[width=0.85\linewidth]{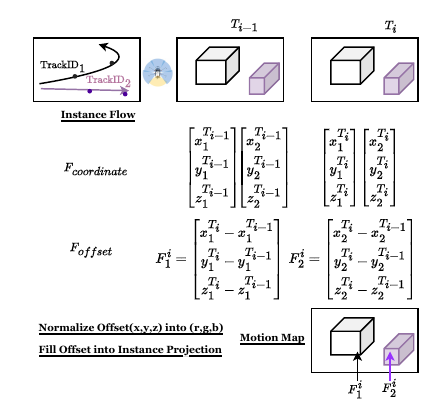} 
\caption{Illustration of the extraction process for motion map in Instance Flow Guider. 
We calculate the motion vector $F_{\text{offset}}$ for each instance, and render the projected box using $F_{\text{offset}}$. 
}
\label{fig: traj}
\vspace{-0.5cm}
\end{figure}

\subsection{Instance Flow Guider}
\label{section: dynamic}


To ensure instance-level temporal consistency, we introduce a lightweight \textbf{I}nstance \textbf{F}low \textbf{G}uider (\textbf{IFG}) module, which enables the model to track, retrieve, and propagate instance features over time, preserving instance attributes such as color and texture. Fig.~\ref{fig: traj} provides an overview of Instance Flow Guider module.


\noindent\textbf{Tracking Instance Position.}
We leverage track IDs to  capture motion trajectories of surrounding instances, termed as \textit{instance flow}, which supports both continuous and non-continuous trajectories:
\begin{equation}
F_{i} = \left \{ (x_{i}^{t}, y_{i}^{t}, z_{i}^{t}, v_{i}^{t}) \right \}_{t=0}^{T-1},
\end{equation}
\noindent where \((x_{j}^{i}, y_{j}^{i}, z_{j}^{i})\) denotes the position of instance \(i\) at frame \(t\), and \( v_i^t \in \{0, 1\} \) indicates whether instance \( i \) is visible at frame \( t \).
We define the most recent visible frame as:
\begin{equation}
\tau(i,t) = \max\{t' \mid t' < t,\ v_i^{t'} = 1\}.
\end{equation}
If $i$ was never visible before $t$, $\tau(i,t)$ is undefined.
\noindent To encode the positional displacement of an instance from previous frame to current frame, we define \textit{instance flow offset}:  
$$
F_{i_{\text{offset}}}^t =
\begin{cases}
(x_i^t - x_i^{\tau(i,t)},\ y_i^t - y_i^{\tau(i,t)},\ z_i^t - z_i^{\tau(i,t)}), \\
\hspace*{5em} \text{if } v_i^t = 1 \text{ and } \tau(i,t) \text{ exists} \\
(0,\ 0,\ 0),\quad \text{otherwise}
\end{cases}
$$
\noindent which propagates features from $\tau(i,t)$ to $t$, effectively reassociating reappearing instances with their pre-occlusion states. 
The full offset map for all \( N \) instances at frame \( t \) is:
\begin{equation}
F_{\text{offset}}^{t} = \left\{ F_{i_{\text{offset}}}^t \right\}_{i=0}^{N-1},
\end{equation}
which serves as a motion condition for tracking the corresponding instance’s position in the previous frame.

\noindent \textbf{Visualizing Motion Conditions.} 
To ensure alignment between the motion condition and the corresponding instance's geometric position,  
we transform \( F_{\text{offset}}^{t} \) into a motion map \( h^t \in \mathbb{R}^{H \times W \times 3} \), which belongs to the same latent space of video patches.  
The 3D bounding box of instance \( i \) is projected onto the camera’s First-Person View to obtain its 2D projection region.  
Each pixel within this region at frame \( t \) is assigned the positional displacement of the corresponding instance:  
\begin{equation}
h^t(h, w) = 
\begin{cases}
F_{i_{\text{offset}}}^t, & \text{if instance } i \text{ projects onto } \\
                             &  (h, w) \text{ at frame } t, \\
0, & \text{otherwise.}
\end{cases}
\end{equation}
\noindent The positional displacements of all instances \( F_{\text{offset}}^{t} \) are collectively encoded into the motion map \( h^t \).  
Notably, the first frame employs a fully-zero map: \( h^{t=0} = 0 \).

\noindent Afterward, to visualize the motion map \( h \in \mathbb{R}^{T \times H \times W \times 3} \), we transform \( h \) into the RGB color space, generating \( h^{vis} \in \mathbb{R}^{T \times H \times W \times 3} \) through a flow visualization technique: \(x\)-offset is mapped to the red (\(R\)) channel, \(y\)-offset is mapped to the green (\(G\)) channel, and \(z\)-offset is mapped to the blue (\(B\)) channel.  
To efficiently encode motion information, we leverage the same VAE encoder \( E_{\text{vae}} \) to compress the motion maps \( h^{vis} \), achieving an 8× spatial downsampling, yielding a compact motion latent representation: $h_{m} \in \mathbb{R}^{T \times h \times w \times 4}$.

\noindent \textbf{Retrieving and Propagating Instance Features.}
Once the motion vector is encoded, we fuse instance motion condition features with the corresponding base block outputs using ControlNet~\cite{zhang2023adding}.  
These fused representations are then passed through temporal attention layers in ST-DiT blocks.
The encoded motion map, which comprises instance flow and instance mask, enables the temporal attention layers to selectively incorporate instance information from adjacent frames.
As a result, the semantic attributes of each instance (e.g., instance category, color, and texture) are retrieved from the past frame and efficiently propagated forward.
By conditioning the ST-DiT model \( D_{\theta} \) on both the spatial location and feature tokens of each instance, we explicitly control where and how each instance should appear in the generated video.
This approach significantly enhances instance-level temporal consistency in driving video synthesis, effectively addressing issues such as color shifts and texture inconsistencies. 

Unlike methods such as GEM \cite{hassan2024gemgeneralizableegovisionmultimodal}, which rely on external feature-extraction models (e.g., DINOv2 \cite{oquab2024dinov2learningrobustvisual}) to obtain instance features, IFG directly derives instance motion and semantic attributes from the generated video itself. This eliminates the need for additional deep feature extractors, making IFG computationally efficient while still ensuring strong instance-level temporal consistency.

\subsection{Spatial Geometric Aligner}
\label{section: static}
To ensure spatial geometric fidelity, we introduce a lightweight \textbf{S}patial \textbf{G}eometric \textbf{A}ligner (\textbf{SGA}) module, which enables the model accurately capture instance spatial locations and occlusion hierarchy. Fig.~\ref{fig: main}(c) provides an overview of the SGA module.
 

To enable accurate spatial localization, we transform 3D bounding boxes into the camera's perspective view, extracting instance bounding box projections as control elements. This transformation uses the camera's intrinsic and extrinsic parameters, ensuring spatial consistency between the world coordinate system and the image plane.

A 3D bounding box $ (c, b) $ consists of a class label $ c $ and box position $ b $. The 3D coordinates $ b_w $ in the world coordinate system, representing one of the eight corner points of $ b $, are first converted into the ego vehicle coordinate system using the inverse ego rotation $ R_e $ and translation $ T_e $:
\begin{equation}
    b_e = R_e^{-1} (b_w - T_e).
\end{equation}
Next, the coordinates are transformed into the camera coordinate system using the calibrated sensor extrinsic matrix:
\begin{equation}
b_c = R_s^{-1} (b_e - T_s),
\end{equation}
where $ R_s $ and $ T_s $ represent the camera's rotation and translation relative to the vehicle. The 3D points $ b_c = [x_c, y_c, z_c] $ are then projected onto the 2D image plane using the camera's intrinsic matrix $ K $.
The 2D pixel coordinates are computed as:
\begin{equation}
\begin{bmatrix} u & v & - \end{bmatrix}^\top = K \begin{bmatrix} x_c / z_c & y_c / z_c & 1 \end{bmatrix}^\top.
\end{equation}
The resulting 2D pixel coordinates render instance-aware spatial constraints in the RGB domain, with distinct colors representing different classes $ c $. These constraints are encoded by the $ E_{\text{layout}} $ encoder into the projected bounding box condition embedding $ h_{\text{box}} $. This explicit transformation pipeline ensures precise instance localization.

To establish a consistent occlusion hierarchy, we explicitly resolve the relative depth order of instances. During the transformation from the camera coordinate system to the image plane, the depth component $ z_c $ represents the distance from an instance's corner point to the camera optical center along the optical axis. We use this depth information as a control condition, helping the model understand occlusion relationships and ensuring closer instances correctly occlude farther ones. 

For each instance, the 3D bounding box is defined by eight corner points in the camera coordinate system, denoted as $ b^i_t \in \mathbb{R}^{8 \times 3} $, where each row $ b_c = [u, v, z_c] $ encodes the position and depth of a corner point in the 2D image plane. We apply Fourier embedding to each corner point and pass it through an MLP to obtain the depth order representation:  
\begin{align}
    h^{b_i}_t &= \operatorname{MLP}_p(\operatorname{Fourier}(b^i_t)) = E_{depth}(b^i_t).
    \label{equ:box_pos}
\end{align}
The hidden states for all bounding boxes in frame $ t $ are represented as $ h_{depth} = [h^{b_1}_t, \dots, h^{b_{N_{t}}}_t] $, where $ N_{t} $ is the number of instances. To integrate depth-based spatial information, we fuse $ h_{depth} $ with the projected bounding box condition embedding $ h_{\text{box}} $ using cross-attention:  
\begin{align}
    h_{\text{vehicle}} =  \text{CrossAttn}(h_{\text{box}}, h_{depth}).
\end{align}
This enables the model to effectively capture spatial locations and occlusion relationships, ensuring geometric fidelity in generated driving videos.

\subsection{CARLA based Procedural Scenario Simulation} 
\label{section:carla}
We used CARLA's autopilot and map system to define obstacle interaction behaviors (e.g., cut-in, braking) and extracted intermediate 3D bounding boxes, lane markings, and drivable areas. Leveraging nuScenes' ego vehicle coordinate system and camera parameters, we projected this data into multiple viewpoints, converting it into model control conditions. This enabled scene generation. The system efficiently utilizes CARLA's waypoint mechanism to randomly generate diverse events across maps, providing a rich set of control conditions.
\section{Experiment}
\label{section: exp}

\begin{figure*}[t]
\centering
\includegraphics[width=0.9\linewidth]{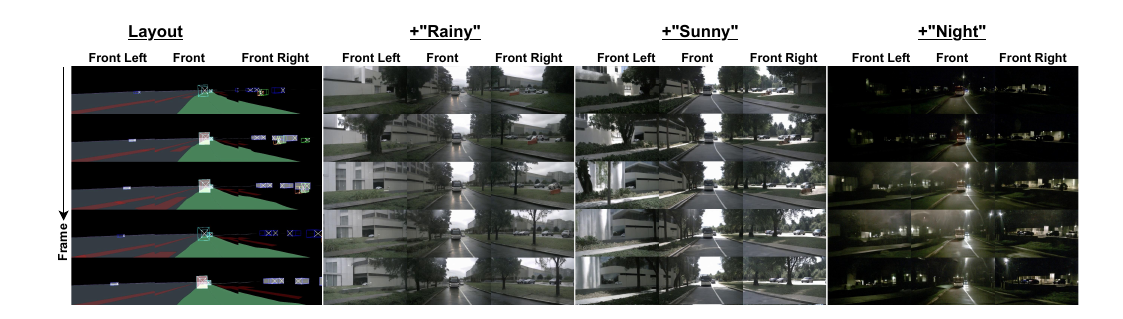} 
\caption{\textbf{Text control}. By adding "Rainy," "Sunny," and "Night" to the original text prompt, while keeping other conditions unchanged, our model represents strong ability to edit videos effectively. 
}
\label{fig:edit}
\end{figure*}

\begin{figure*}[tb]
\centering
\includegraphics[width=0.76\linewidth]{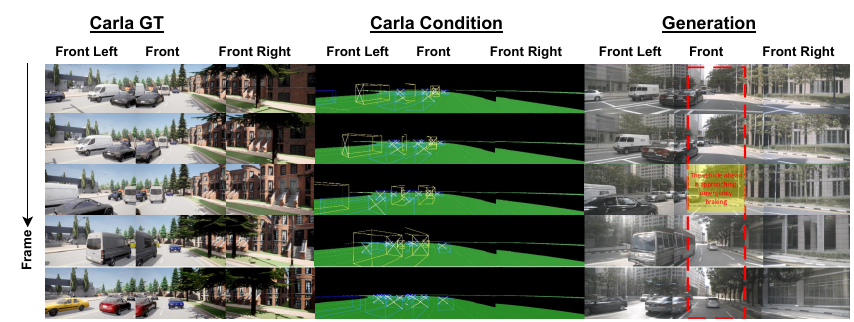} 
\caption{\textbf{Long-tail scenarios simulation.} 
Using CARLA’s highly configurable simulation environment, we create synthetic control conditions which represent complex driving scenarios (e.g., sudden braking), and then utilize InstaDrive to generate corresponding videos.
}
\label{fig:carla}
\end{figure*}

\subsection{Setups}
\noindent \textbf{Datasets and Baselines.} 
We train and evaluate our model on the nuScenes dataset \cite{nuScenes}. To benchmark our approach, we compare it with state-of-the-art driving world models, including BEVControl \cite{bevcontrol}, DriveDiffusion \cite{li2023drivingdiffusionlayoutguidedmultiviewdriving}, DriveDreamer2 \cite{drivedreamer2}, Panacea \cite{wen2024panacea}, and MagicDrive-V2 \cite{gao2024magicdrivedithighresolutionlongvideo}.

\noindent \textbf{Metrics.}
For realism assessment, we use FID \cite{FID} and FVD \cite{Thomas2018fvd} to measure video quality. To evaluate instance-level temporal consistency, we test our model in real-world autonomous driving scenarios using the multi-object tracking (MOT) task. MOT ensures consistent object tracking across frames, minimizing ID switches (IDSW) and drift. Following Panacea \cite{wen2024panacea}, we employ the StreamPETR model \cite{streampetr} and evaluate performance using standard MOT metrics: AMOTA, AMOTP, RECALL, MOTA, and IDS. Additionally, we evaluate spatial geometric fidelity by measuring the alignment between generated videos and their conditioned sequences, ensuring accurate preservation of geometric structures. This is assessed through perception tasks, as precise detection and localization are fundamental to perception. Thus, perception performance directly indicates the accuracy of object localization and occlusion relationships. Using the StreamPETR model \cite{streampetr}, we employ metrics such as the nuScenes Detection Score (NDS), mean Average Precision (mAP), mean Average Orientation Error (mAOE), and mean Average Velocity Error (mAVE). We note that Bevfusion \cite{liu2023bevfusion} has also been used for perception metric evaluation \cite{gao2023magicdrive}, but it is based on an image model, performs worse than the video-based model StreamPETR, and cannot provide object tracking metrics. Therefore, we still choose StreamPETR for evaluation.

\noindent For metrics and results on the planning task using pretrained UniAD \cite{shi2016uniad} model, please refer to Appendix.

\begin{table}[tb]
    \centering
    \footnotesize
        \centering
        \resizebox{0.475\textwidth}{!}{
        \setlength{\tabcolsep}{8pt}
        \begin{tabular}{lcccc}
            \toprule
            Method     &Multi-View &Multi-Frame  & FVD$\downarrow$ & FID$\downarrow$ \\
            \hline
            BEVControl \cite{bevcontrol}  &$\checkmark$   &  & - & 24.85 \\
            DrivingDiffusion \cite{li2023drivingdiffusionlayoutguidedmultiviewdriving}   &$\checkmark$ &$\checkmark$  & 332 & 15.83\\
            Panacea \cite{wen2024panacea}   & $\checkmark$ &$\checkmark$  & 139 & 16.96\\  
            MagicDrive-V2 \cite{gao2024magicdrivedithighresolutionlongvideo}   &$\checkmark$ &$\checkmark$  & 94.84 & 20.91\\
            DriveScape \cite{drivescape}  & 
            $\checkmark$ &$\checkmark$  & 76.39 & 8.34\\
            DriveDreamer2 \cite{drivedreamer2}   &$\checkmark$ &$\checkmark$  & 55.7 & 11.2\\
            \hline
            \rowcolor[gray]{.9} 
            \titleDef~   &$\checkmark$  &$\checkmark$  & \textcolor{blue}{38.06}  & \textcolor{blue}{3.96}  \\
            \bottomrule
        \end{tabular}
        }
        \caption{Comparing with SoTA methods on the validation set of the nuScenes dataset. We generate the entire validation set without applying any post-processing strategies to select specific samples.}
        \label{tab:fvd}
\end{table}

\subsection{Training Details}
Our method is implemented based on OpenSora \cite{opensora}. 
All training inputs were set to 16x256x448 and conducted on 8 A100 GPUs.
Experimental results show that our method can stably generate over 200 frames. For more implementation details, please refer to Appendix.

\subsection{Main Results}

\subsubsection{Quantitative Analysis}
To verify the high fidelity of our generated videos, we compare our approach with various state-of-the-art driving world models. 
We generate training and validation data using the nuScenes dataset’s labels as conditions. 
For fairness, we generate the entire dataset without applying any post-processing strategies to select specific samples.

\noindent\textbf{Realism of Images.} 
Our generated videos exhibit higher visual quality, achieving an FID of $3.96$, as shown in Tab.~\ref{tab:fvd}. It substantially outperforms those of all counterparts, including both video-based methods like DriveDreamer2 ~\cite{drivedreamer2} and image-based solutions such as BEVControl ~\cite{bevcontrol}. 

\begin{table}[tb]
    \centering
    \footnotesize
    \resizebox{0.47\textwidth}{!}{
    \setlength{\tabcolsep}{4pt}
    \begin{tabular}{lcccccc}
        \toprule
        Method & Real & Generated & AMOTA$\uparrow$ & AMOTP$\downarrow$ & IDS$\downarrow$  \\
        \hline
        Oracle & $\checkmark$  & - & 0.289 & 1.419 & 687 \\
        DriveDreamer2 & $\checkmark$  & $\checkmark$   & 0.313   & 1.387 & 593 \textcolor{blue}{(-94)}  \\
        \rowcolor[gray]{.9} 
        \titleDef~ (Ours) & $\checkmark$  & $\checkmark$   & 0.496    & 1.376 & 532 \textcolor{blue}{(-155)}\\
        \bottomrule
    \end{tabular}
    }
    \caption{Comparison on multi-object tracking task with MagicDrive-V2 \cite{gao2024magicdrivedithighresolutionlongvideo} based on a pre-trained StreamPETR model.
    }
    \label{tab:tracking}
\end{table}

\noindent\textbf{Instance-Level Temporal Consistency.} 
Our method significantly reduces FVD to $38.06$ in Tab.~\ref{tab:fvd}, due to the Instance Flow Guider module preserving instance attributes over frames and therefore enhancing instance-level temporal consistency.
Additionally, we also assess our model in real-world autonomous driving applications using the multi-object tracking (MOT) task in Tab.~\ref{tab:tracking}, 
as MOT requires consistent tracking of the same object across frames while minimizing ID switches, making it a strong indicator of instance-level temporal consistency.
Specifically, we generate data using the nuScenes validation set’s labels as conditions. 
We then re-train the object tracking model StreamPETR \cite{streampetr} by integrating the generated data with real data.
The MOT model's performance improves significantly, achieving a lower IDS of $532$ compared to the originally pre-trained StreamPETR, indicating 
our framework’s effectiveness in producing instance-level temporally consistent synthetic data.

\begin{table}[htbp]
    \centering
    \footnotesize
        \centering
        \resizebox{0.36\textwidth}{!}{
        \setlength{\tabcolsep}{7pt}
        \begin{tabular}{cccc}
            \toprule
              Method   & Real & Generated &  NDS$\uparrow$\\
            \hline
                Oracle & $\checkmark$ & -  &46.90 \\ 
                Panacea & - &$\checkmark$ &32.10 (68.00\%) \\
                MagicDrive-V2 & - &$\checkmark$ &36.82 (78.51\%) \\
               \cellcolor[gray]{.9} \titleDef~     &\cellcolor[gray]{.9} - 
               &\cellcolor[gray]{.9}  $ \cellcolor[gray]{.9} \checkmark$ 
               & \cellcolor[gray]{.9} \textcolor{blue}{40.51 (86.38\%) }
               \\
            \bottomrule
        \end{tabular}
        }
         \caption{
         Comparison on perception task using the generated nuScenes validation set in (T+I)2V scenarios.
We use pre-trained perception model StreamPETR \cite{streampetr} to evaluate. 
Our model outperformed most baseline models across the board without any post-refine process, underscoring a better capability of spatial localization and occlusion hierarchy understanding.
         }
        \label{tab:perception}
\end{table}

\noindent\textbf{Spatial Geometric Fidelity.} 
\noindent \textit{Data Augmentation Performance.} 
We assess our model in real-world autonomous driving applications using the perception task, 
as the perception task fundamentally relies on precise detection and localization. Thus, perception performance directly reflects the accuracy of object localization and occlusion relationships. 
In Tab.~\ref{tab:perception_augment}, 
we first train StreamPETR exclusively on our generated training dataset, and it achieves a mAP of $35.5\%$, equivalent to $92.69\%$ of the performance obtained by models trained solely on the real nuScenes training dataset. 
These results highlight that the generated dataset is not only a viable substitute for real data but also highly effective in training perception models independently.
Moreover, we re-train StreamPETR by integrating the generated data with real data; the perception model's performance improves significantly, achieving a NDS of $51.9$, marking a $3.6$-point increase over the model trained exclusively on real data. This underscores the substantial value of incorporating generated data into the training pipeline.



\noindent \textbf{Perception Validation Performance.}  
Additionally, we use the pre-trained StreamPETR model to evaluate the generated validation set of the nuScenes.
In Tab.~\ref{tab:perception}, our model achieves a relative performance of $86.38\%$ on the nuScenes Detection Score (NDS), underscoring a better alignment with the control conditions. Additional results on planning task can refer to Appendix.

\subsubsection{Qualitative Analysis}
We conduct a qualitative comparison of \titleDef~ with other SOTA models using the generated videos.

\noindent \textbf{Instance-Level Temporal Consistency.} 
In Fig.\ref{fig:intro}(a), MagicDrive-V2 exhibits an inconsistency where the car's color changes over time. In contrast, our model maintains the car's attributes consistently across all frames, showcasing superior instance-level temporal consistency.

\noindent \textbf{Spatial Localization.} 
In Fig.\ref{fig:intro}(b), MagicDrive-V2 fails to adhere to the bounding box control signal, resulting in spatial deviation. In contrast, our model ensures precise spatial localization, accurately following the control signal.

\noindent \textbf{Occlusion Hierarchy.} 
In Fig.\ref{fig:intro}(c), for the video generated by Panacea, 
the distant bus incorrectly occludes the nearby pedestrian, violating natural occlusion rules. 
In contrast, our model accurately preserves the occlusion hierarchy, ensuring a realistic representation of scene depth.

\subsubsection{Long-tail Scenarios Simulation.}
Our method can simulate diverse long-tail driving scenarios. By modifying text prompts, we can change the weather and time of scenes, as shown in Fig.~\ref{fig:edit}. We can also generate critical long-tail events, such as sudden braking and lane cutting, using control conditions provided by the \textbf{Carla} simulator \cite{Dosovitskiy17}. Carla supplies conditions like 3D bounding box projections, lane line projections, and text descriptions for scenes. Leveraging Carla's highly configurable environment, we create synthetic control conditions for complex and diverse scenarios, such as multi-vehicle intersections, narrow streets, or sudden obstacles, which are challenging to capture in real-world data. In Fig.~\ref{fig:carla}, we demonstrate our model's ability to generate long-tail videos corresponding to these conditions. The results show that our method not only replicates realistic conditions but also seamlessly adapts to complex scenarios generated by Carla, enhancing its applicability in autonomous driving research. For more visualization details, please refer to the Appendix.

\begin{table}[htbp]
    \centering
    \footnotesize
    \resizebox{0.475\textwidth}{!}{
    \setlength{\tabcolsep}{4pt}
    \begin{tabular}{lcccccc}
        \toprule
        Method & Real & Gen. & mAP$\uparrow$ & mAOE$\downarrow$ & mAVE$\downarrow$ & NDS$\uparrow$ \\
        \hline
        Panacea & $\checkmark$  & - & 34.5 & 59.4 & 29.1 & 46.9 \\
        Panacea & - &$\checkmark$  & 22.5 & 72.7 & 46.9 & 36.1 \\
        Panacea & $\checkmark$  & $\checkmark$   & 37.1 \textcolor{blue}{(+2.6\%)} & 54.2 & 27.3 & 49.2 \textcolor{blue}{(+2.3\%)} \\
        \hline
        \titleDef~ (Re-Impl.) &$\checkmark$  & -  & 38.3 & 62.1 & 28.8 & 48.3 \\
        \titleDef~ (Ours) & -  &$\checkmark$  & 35.5 & 59.7 & 29.4 & 43.67 \\
        \rowcolor[gray]{.9} 
        \titleDef~ (Ours) & $\checkmark$  & $\checkmark$   & 42.0 \textcolor{blue}{(+3.7\%)} & 53.2 & 26.8 & 51.9 \textcolor{blue}{(+3.6\%)} \\
        \bottomrule
    \end{tabular}
    }
    \caption{Comparison on perception task with Panacea. We involve data augmentation using synthetic training data to train StreamPETR. The perception model's performance improves significantly, showing the substantial value of incorporating generated data into the training pipeline. 
    }
    \label{tab:perception_augment}
\end{table}

\begin{table}[htbp]
    \centering
    \footnotesize
        \centering
        \resizebox{0.475\textwidth}{!}{
        \setlength{\tabcolsep}{4pt}
        \begin{tabular}{lcccc}
            \toprule
            Settings        & FVD$\downarrow$ & FID$\downarrow$ & NDS$\uparrow$ & IDS$\downarrow$ \\
            \hline
            \rowcolor[gray]{.9} 
            \titleDef~   & 38.06 & 3.96 & 40.51 & 532\\
            \hline
            w/o IFG    & 58.86 \textcolor{blue}{(+20.8)} & 6.28 \textcolor{blue}{(+2.32)} & 40.23 \textcolor{blue}{(-0.28)} & 1189 \textcolor{blue}{(+653)}\\ 
            w/o SGA & 41.21 \textcolor{blue}{(+3.15)}& 4.22 \textcolor{blue}{(+0.26)} & 37.31 \textcolor{blue}{(-3.20)} & 723 \textcolor{blue}{(+187)}\\    
            w/o SGA(Box) & 39.27 \textcolor{blue}{(+1.21)}& 4.07 \textcolor{blue}{(+0.11)} & 38.22 \textcolor{blue}{(-2.29)} & 628 \textcolor{blue}{(+92)}\\ 
            w/o SGA(Depth)   & 40.46 \textcolor{blue}{(+2.40)} & 4.15 \textcolor{blue}{(+0.19)} & 39.13 \textcolor{blue}{(-1.38)} & 669 \textcolor{blue}{(+133)}\\
            \bottomrule
        \end{tabular}
        }
        \caption{Ablation study results in (T+I)2V scenarios on the generated nuScenes validation set.}
        \label{tab: ablation}
\end{table}

\begin{figure}[htbp]
\centering
\includegraphics[width=1.0\linewidth]{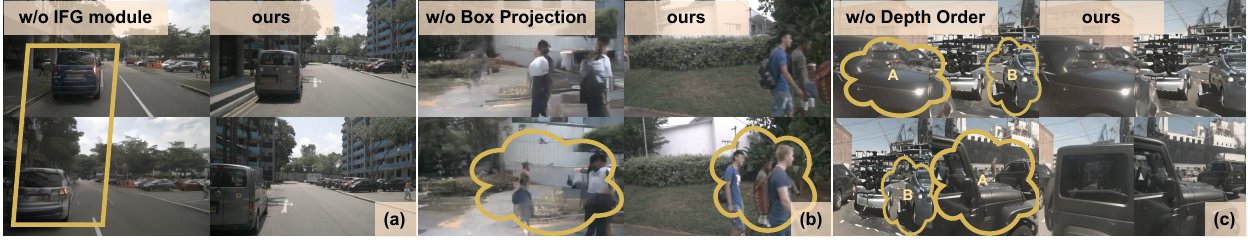} 
\caption{
Ablation study of two key modules. \textbf{Zoom for better view.}
\textbf{(a)} Removing the IFG causes the car's color to change over time, breaking temporal consistency.
\textbf{(b)} Replacing box projection with box coordinates in the Spatial Geometric Aligner module causes misalignment between the passenger and the control box, resulting in failed spatial localization.
\textbf{(c)} Removing depth cues in the Spatial Geometric Aligner results in the nearby car A failing to occlude the distant car B, disrupting occlusion hierarchy.
}
\label{fig:ablation_compare}
\vspace{-0.5cm}
\end{figure}

\subsection{Ablation Study}
We validate two key modules through qualitative and quantitative analyses, demonstrating their effectiveness and robustness.
The qualitative comparison is in Fig. \ref{fig:ablation_compare}. 


\noindent \textbf{Instance Flow Guider.}
To evaluate the impact of the IFG module, we conduct an ablation study by removing the instance flow injection. 
In Tab.~\ref{tab: ablation}, the absence of instance flow leads to a significant drop of $20.8$ in FVD and an increase of $653$ in IDS (ID Switch) for the multi-object tracking task.
These results underscore the critical role of the IFG module in preserving instance-level temporal consistency.

\noindent \textbf{Spatial Geometric Aligner.}
To evaluate the influence of the SGA module, we remove the injection of depth cues and replace the box projection with box coordinates as the control signal, seperately. 
As shown in Tab.~\ref{tab: ablation}, the absence of SGA module results in a significant degradation of \textbf{3.20} in NDS, highlighting its critical role in understanding spatial localization and occlusion hierarchy.

\noindent Our method can also generate videos without the initial frame as condition. To evaluate the impact of two key modules in the T2V scenario, please refer to Appendix.

\section{Conclusion}
\label{section: conclusion}
We propose \titleDef, a driving world model that enhances instance-level temporal consistency and spatial geometric fidelity. Our approach introduces two key advancements: the Instance Flow Guider module, which extracts and propagates instance features across frames to preserve instance identity over time, and the Spatial Geometric Aligner module, which ensures precise instance positioning and explicitly models occlusion hierarchies. By incorporating these instance-aware mechanisms, \titleDef~achieves SOTA generation quality and significantly improves downstream autonomous driving tasks. Finally, we leverage CARLA’s autopilot to generate rare yet safety-critical driving scenarios, demonstrating the immense potential of \titleDef~ in long-tail simulation.


\section*{Acknowledgements}
This work was supported by the National Natural Science Foundation of China (No. 62332016) and the Key Research Program of Frontier Sciences, CAS (No. ZDBS-LY-JSC001).



{
    \small
    \bibliographystyle{ieeenat_fullname}
    \bibliography{main}

\begin{thebibliography}{60}
\providecommand{\natexlab}[1]{#1}
\providecommand{\url}[1]{\texttt{#1}}
\expandafter\ifx\csname urlstyle\endcsname\relax
  \providecommand{\doi}[1]{doi: #1}\else
  \providecommand{\doi}{doi: \begingroup \urlstyle{rm}\Url}\fi

\bibitem[An et~al.(2023)An, Zhang, Yang, Gupta, Huang, Luo, and Yin]{an2023latent}
Jie An, Songyang Zhang, Harry Yang, Sonal Gupta, Jia-Bin Huang, Jiebo Luo, and Xi Yin.
\newblock Latent-shift: Latent diffusion with temporal shift for efficient text-to-video generation.
\newblock \emph{arXiv preprint arXiv:2304.08477}, 2023.

\bibitem[Bar-Tal et~al.(2024)Bar-Tal, Chefer, Tov, Herrmann, Paiss, Zada, Ephrat, Hur, Liu, Raj, et~al.]{bar2024lumiere}
Omer Bar-Tal, Hila Chefer, Omer Tov, Charles Herrmann, Roni Paiss, Shiran Zada, Ariel Ephrat, Junhwa Hur, Guanghui Liu, Amit Raj, et~al.
\newblock Lumiere: A space-time diffusion model for video generation.
\newblock In \emph{SIGGRAPH Asia 2024 Conference Papers}, pages 1--11, 2024.

\bibitem[Blattmann et~al.(2023{\natexlab{a}})Blattmann, Dockhorn, Kulal, Mendelevitch, Kilian, Lorenz, Levi, English, Voleti, Letts, et~al.]{svd}
Andreas Blattmann, Tim Dockhorn, Sumith Kulal, Daniel Mendelevitch, Maciej Kilian, Dominik Lorenz, Yam Levi, Zion English, Vikram Voleti, Adam Letts, et~al.
\newblock Stable video diffusion: Scaling latent video diffusion models to large datasets.
\newblock \emph{arXiv preprint arXiv:2311.15127}, 2023{\natexlab{a}}.

\bibitem[Blattmann et~al.(2023{\natexlab{b}})Blattmann, Rombach, Ling, Dockhorn, Kim, Fidler, and Kreis]{blattmann2023align}
Andreas Blattmann, Robin Rombach, Huan Ling, Tim Dockhorn, Seung~Wook Kim, Sanja Fidler, and Karsten Kreis.
\newblock Align your latents: High-resolution video synthesis with latent diffusion models.
\newblock In \emph{Proceedings of the IEEE/CVF conference on computer vision and pattern recognition}, pages 22563--22575, 2023{\natexlab{b}}.

\bibitem[Cabon et~al.(2020)Cabon, Murray, and Humenberger]{cabon2020virtual}
Yohann Cabon, Naila Murray, and Martin Humenberger.
\newblock Virtual kitti 2.
\newblock \emph{arXiv preprint arXiv:2001.10773}, 2020.

\bibitem[Caesar et~al.(2020)Caesar, Bankiti, Lang, Vora, Liong, Xu, Krishnan, Pan, Baldan, and Beijbom]{nuScenes}
Holger Caesar, Varun Bankiti, Alex~H Lang, Sourabh Vora, Venice~Erin Liong, Qiang Xu, Anush Krishnan, Yu Pan, Giancarlo Baldan, and Oscar Beijbom.
\newblock nuscenes: A multimodal dataset for autonomous driving.
\newblock In \emph{Proceedings of the IEEE/CVF conference on computer vision and pattern recognition}, pages 11621--11631, 2020.

\bibitem[Chen et~al.(2024)Chen, Wu, Chitta, Jaeger, Geiger, and Li]{chen2024end}
Li Chen, Penghao Wu, Kashyap Chitta, Bernhard Jaeger, Andreas Geiger, and Hongyang Li.
\newblock End-to-end autonomous driving: Challenges and frontiers.
\newblock \emph{IEEE Transactions on Pattern Analysis and Machine Intelligence}, 2024.

\bibitem[Dosovitskiy et~al.(2017)Dosovitskiy, Ros, Codevilla, Lopez, and Koltun]{Dosovitskiy17}
Alexey Dosovitskiy, German Ros, Felipe Codevilla, Antonio Lopez, and Vladlen Koltun.
\newblock {CARLA}: {An} open urban driving simulator.
\newblock In \emph{Proceedings of the 1st Annual Conference on Robot Learning}, pages 1--16, 2017.

\bibitem[Gao et~al.(2024{\natexlab{a}})Gao, Shi, Zhang, Wang, Xiao, and Chen]{gao2024ca2}
Kaifeng Gao, Jiaxin Shi, Hanwang Zhang, Chunping Wang, Jun Xiao, and Long Chen.
\newblock Ca2-vdm: Efficient autoregressive video diffusion model with causal generation and cache sharing.
\newblock \emph{arXiv preprint arXiv:2411.16375}, 2024{\natexlab{a}}.

\bibitem[Gao et~al.(2024{\natexlab{b}})Gao, Chen, Xiao, Hong, Li, and Xu]{gao2024magicdrivedithighresolutionlongvideo}
Ruiyuan Gao, Kai Chen, Bo Xiao, Lanqing Hong, Zhenguo Li, and Qiang Xu.
\newblock Magicdrivedit: High-resolution long video generation for autonomous driving with adaptive control, 2024{\natexlab{b}}.

\bibitem[Gao et~al.(2024{\natexlab{c}})Gao, Chen, Xie, Lanqing, Li, Yeung, and Xu]{gao2023magicdrive}
Ruiyuan Gao, Kai Chen, Enze Xie, Hong Lanqing, Zhenguo Li, Dit-Yan Yeung, and Qiang Xu.
\newblock Magicdrive: Street view generation with diverse 3d geometry control.
\newblock In \emph{ICLR}, 2024{\natexlab{c}}.

\bibitem[Guo et~al.(2020)Guo, Wang, Yang, Lv, Liu, Wu, and Huang]{guo2020gan}
Xi Guo, Zhicheng Wang, Qin Yang, Weifeng Lv, Xianglong Liu, Qiong Wu, and Jian Huang.
\newblock Gan-based virtual-to-real image translation for urban scene semantic segmentation.
\newblock \emph{Neurocomputing}, 394:\penalty0 127--135, 2020.

\bibitem[Guo et~al.(2022)Guo, Wu, Wang, Su, Su, Gan, Huang, and Yang]{guo2022learning}
Xi Guo, Wei Wu, Dongliang Wang, Jing Su, Haisheng Su, Weihao Gan, Jian Huang, and Qin Yang.
\newblock Learning video representations of human motion from synthetic data.
\newblock In \emph{Proceedings of the IEEE/CVF Conference on Computer Vision and Pattern Recognition}, pages 20197--20207, 2022.

\bibitem[Guo et~al.(2024)Guo, Ding, Dou, Zhang, Tang, and Wu]{guo2024infinitydrive}
Xi Guo, Chenjing Ding, Haoxuan Dou, Xin Zhang, Weixuan Tang, and Wei Wu.
\newblock Infinitydrive: Breaking time limits in driving world models.
\newblock \emph{arXiv preprint arXiv:2412.01522}, 2024.

\bibitem[Guo et~al.(2023)Guo, Yang, Rao, Wang, Qiao, Lin, and Dai]{guo2023animatediff}
Yuwei Guo, Ceyuan Yang, Anyi Rao, Yaohui Wang, Yu Qiao, Dahua Lin, and Bo Dai.
\newblock Animatediff: Animate your personalized text-to-image diffusion models without specific tuning.
\newblock \emph{arXiv preprint arXiv:2307.04725}, 2023.

\bibitem[Gupta et~al.(2024)Gupta, Yu, Sohn, Gu, Hahn, Li, Essa, Jiang, and Lezama]{gupta2024photorealistic}
Agrim Gupta, Lijun Yu, Kihyuk Sohn, Xiuye Gu, Meera Hahn, Fei-Fei Li, Irfan Essa, Lu Jiang, and Jos{\'e} Lezama.
\newblock Photorealistic video generation with diffusion models.
\newblock In \emph{European Conference on Computer Vision}, pages 393--411. Springer, 2024.

\bibitem[HaCohen et~al.(2024)HaCohen, Chiprut, Brazowski, Shalem, Moshe, Richardson, Levin, Shiran, Zabari, Gordon, et~al.]{hacohen2024ltx}
Yoav HaCohen, Nisan Chiprut, Benny Brazowski, Daniel Shalem, Dudu Moshe, Eitan Richardson, Eran Levin, Guy Shiran, Nir Zabari, Ori Gordon, et~al.
\newblock Ltx-video: Realtime video latent diffusion.
\newblock \emph{arXiv preprint arXiv:2501.00103}, 2024.

\bibitem[Hassan et~al.(2024)Hassan, Stapf, Rahimi, Rezende, Haghighi, Brüggemann, Katircioglu, Zhang, Chen, Saha, Cannici, Aljalbout, Ye, Wang, Davtyan, Salzmann, Scaramuzza, Pollefeys, Favaro, and Alahi]{hassan2024gemgeneralizableegovisionmultimodal}
Mariam Hassan, Sebastian Stapf, Ahmad Rahimi, Pedro M~B Rezende, Yasaman Haghighi, David Brüggemann, Isinsu Katircioglu, Lin Zhang, Xiaoran Chen, Suman Saha, Marco Cannici, Elie Aljalbout, Botao Ye, Xi Wang, Aram Davtyan, Mathieu Salzmann, Davide Scaramuzza, Marc Pollefeys, Paolo Favaro, and Alexandre Alahi.
\newblock Gem: A generalizable ego-vision multimodal world model for fine-grained ego-motion, object dynamics, and scene composition control, 2024.

\bibitem[He et~al.(2022)He, Yang, Zhang, Shan, and Chen]{he2022latent}
Yingqing He, Tianyu Yang, Yong Zhang, Ying Shan, and Qifeng Chen.
\newblock Latent video diffusion models for high-fidelity video generation with arbitrary lengths.
\newblock \emph{arXiv preprint arXiv:2211.13221}, 2022.

\bibitem[Heusel et~al.(2017)Heusel, Ramsauer, Unterthiner, Nessler, and Hochreiter]{FID}
Martin Heusel, Hubert Ramsauer, Thomas Unterthiner, Bernhard Nessler, and Sepp Hochreiter.
\newblock Gans trained by a two time-scale update rule converge to a local nash equilibrium.
\newblock In \emph{Proceedings of the 31st International Conference on Neural Information Processing Systems}, page 6629–6640, Red Hook, NY, USA, 2017. Curran Associates Inc.

\bibitem[Ho et~al.(2022{\natexlab{a}})Ho, Chan, Saharia, Whang, Gao, Gritsenko, Kingma, Poole, Norouzi, Fleet, et~al.]{ho2022imagen}
Jonathan Ho, William Chan, Chitwan Saharia, Jay Whang, Ruiqi Gao, Alexey Gritsenko, Diederik~P Kingma, Ben Poole, Mohammad Norouzi, David~J Fleet, et~al.
\newblock Imagen video: High definition video generation with diffusion models.
\newblock \emph{arXiv preprint arXiv:2210.02303}, 2022{\natexlab{a}}.

\bibitem[Ho et~al.(2022{\natexlab{b}})Ho, Salimans, Gritsenko, Chan, Norouzi, and Fleet]{ho2022video}
Jonathan Ho, Tim Salimans, Alexey Gritsenko, William Chan, Mohammad Norouzi, and David~J Fleet.
\newblock Video diffusion models.
\newblock \emph{Advances in Neural Information Processing Systems}, 35:\penalty0 8633--8646, 2022{\natexlab{b}}.

\bibitem[Hu(2024)]{hu2024animate}
Li Hu.
\newblock Animate anyone: Consistent and controllable image-to-video synthesis for character animation.
\newblock In \emph{Proceedings of the IEEE/CVF Conference on Computer Vision and Pattern Recognition}, pages 8153--8163, 2024.

\bibitem[Hu et~al.(2023)Hu, Yang, Chen, Li, Sima, Zhu, Chai, Du, Lin, Wang, Lu, Jia, Liu, Dai, Qiao, and Li]{hu2023planningorientedautonomousdriving}
Yihan Hu, Jiazhi Yang, Li Chen, Keyu Li, Chonghao Sima, Xizhou Zhu, Siqi Chai, Senyao Du, Tianwei Lin, Wenhai Wang, Lewei Lu, Xiaosong Jia, Qiang Liu, Jifeng Dai, Yu Qiao, and Hongyang Li.
\newblock Planning-oriented autonomous driving, 2023.

\bibitem[Jia et~al.(2023)Jia, Mao, Liu, Zhao, Wen, Zhang, Zhang, and Wang]{jia2023adriver}
Fan Jia, Weixin Mao, Yingfei Liu, Yucheng Zhao, Yuqing Wen, Chi Zhang, Xiangyu Zhang, and Tiancai Wang.
\newblock Adriver-i: A general world model for autonomous driving.
\newblock \emph{arXiv preprint arXiv:2311.13549}, 2023.

\bibitem[Jiang et~al.(2023)Jiang, Chen, Xu, Liao, Chen, Zhou, Zhang, Liu, Huang, and Wang]{jiang2023vad}
Bo Jiang, Shaoyu Chen, Qing Xu, Bencheng Liao, Jiajie Chen, Helong Zhou, Qian Zhang, Wenyu Liu, Chang Huang, and Xinggang Wang.
\newblock Vad: Vectorized scene representation for efficient autonomous driving.
\newblock In \emph{Proceedings of the IEEE/CVF International Conference on Computer Vision}, pages 8340--8350, 2023.

\bibitem[Lei et~al.(2023)Lei, Hu, Wang, and Liu]{lei2023pyramidflow}
Jiarui Lei, Xiaobo Hu, Yue Wang, and Dong Liu.
\newblock Pyramidflow: High-resolution defect contrastive localization using pyramid normalizing flow.
\newblock In \emph{Proceedings of the IEEE/CVF conference on computer vision and pattern recognition}, pages 14143--14152, 2023.

\bibitem[Li et~al.(2023)Li, Zhang, and Ye]{li2023drivingdiffusionlayoutguidedmultiviewdriving}
Xiaofan Li, Yifu Zhang, and Xiaoqing Ye.
\newblock Drivingdiffusion: Layout-guided multi-view driving scene video generation with latent diffusion model, 2023.

\bibitem[Liu et~al.(2023)Liu, Tang, Amini, Yang, Mao, Rus, and Han]{liu2023bevfusion}
Zhijian Liu, Haotian Tang, Alexander Amini, Xinyu Yang, Huizi Mao, Daniela~L Rus, and Song Han.
\newblock Bevfusion: Multi-task multi-sensor fusion with unified bird's-eye view representation.
\newblock In \emph{2023 IEEE international conference on robotics and automation (ICRA)}, pages 2774--2781. IEEE, 2023.

\bibitem[Oquab et~al.(2024)Oquab, Darcet, Moutakanni, Vo, Szafraniec, Khalidov, Fernandez, Haziza, Massa, El-Nouby, Assran, Ballas, Galuba, Howes, Huang, Li, Misra, Rabbat, Sharma, Synnaeve, Xu, Jegou, Mairal, Labatut, Joulin, and Bojanowski]{oquab2024dinov2learningrobustvisual}
Maxime Oquab, Timothée Darcet, Théo Moutakanni, Huy Vo, Marc Szafraniec, Vasil Khalidov, Pierre Fernandez, Daniel Haziza, Francisco Massa, Alaaeldin El-Nouby, Mahmoud Assran, Nicolas Ballas, Wojciech Galuba, Russell Howes, Po-Yao Huang, Shang-Wen Li, Ishan Misra, Michael Rabbat, Vasu Sharma, Gabriel Synnaeve, Hu Xu, Hervé Jegou, Julien Mairal, Patrick Labatut, Armand Joulin, and Piotr Bojanowski.
\newblock Dinov2: Learning robust visual features without supervision, 2024.

\bibitem[Raffel et~al.(2020)Raffel, Shazeer, Roberts, Lee, et~al.]{raffel2020exploring}
Colin Raffel, Noam Shazeer, Adam Roberts, Katherine Lee, et~al.
\newblock Exploring the limits of transfer learning with a unified text-to-text transformer.
\newblock \emph{Journal of machine learning research}, 21\penalty0 (140):\penalty0 1--67, 2020.

\bibitem[Ros et~al.(2016)Ros, Sellart, Materzynska, Vazquez, and Lopez]{ros2016synthia}
German Ros, Laura Sellart, Joanna Materzynska, David Vazquez, and Antonio~M Lopez.
\newblock The synthia dataset: A large collection of synthetic images for semantic segmentation of urban scenes.
\newblock In \emph{Proceedings of the IEEE conference on computer vision and pattern recognition}, pages 3234--3243, 2016.

\bibitem[Shi et~al.(2016)Shi, Cui, Dobbie, and Ooi]{shi2016uniad}
Xiaogang Shi, Bin Cui, Gillian Dobbie, and Beng~Chin Ooi.
\newblock Uniad: A unified ad hoc data processing system.
\newblock \emph{ACM Transactions on Database Systems (TODS)}, 42\penalty0 (1):\penalty0 1--42, 2016.

\bibitem[Shrivastava et~al.(2017)Shrivastava, Pfister, Tuzel, Susskind, Wang, and Webb]{shrivastava2017learning}
Ashish Shrivastava, Tomas Pfister, Oncel Tuzel, Joshua Susskind, Wenda Wang, and Russell Webb.
\newblock Learning from simulated and unsupervised images through adversarial training.
\newblock In \emph{Proceedings of the IEEE conference on computer vision and pattern recognition}, pages 2107--2116, 2017.

\bibitem[Singer et~al.(2022)Singer, Polyak, Hayes, Yin, An, Zhang, Hu, Yang, Ashual, Gafni, et~al.]{singer2022make}
Uriel Singer, Adam Polyak, Thomas Hayes, Xi Yin, Jie An, Songyang Zhang, Qiyuan Hu, Harry Yang, Oron Ashual, Oran Gafni, et~al.
\newblock Make-a-video: Text-to-video generation without text-video data.
\newblock \emph{arXiv preprint arXiv:2209.14792}, 2022.

\bibitem[Swerdlow et~al.(2023)Swerdlow, Xu, and Zhou]{swerdlow2023street}
Alexander Swerdlow, Runsheng Xu, and Bolei Zhou.
\newblock Street-view image generation from a bird's-eye view layout.
\newblock \emph{arXiv preprint arXiv:2301.04634}, 2023.

\bibitem[Tang et~al.(2023)Tang, Zhang, Chen, Wang, and Furukawa]{tang2023mvdiffusion}
Shitao Tang, Fuyang Zhang, Jiacheng Chen, Peng Wang, and Yasutaka Furukawa.
\newblock Mvdiffusion: Enabling holistic multi-view image generation with correspondence-aware diffusion.
\newblock \emph{arXiv preprint arXiv:2307.01097}, 2023.

\bibitem[Tobin et~al.(2017)Tobin, Fong, Ray, Schneider, Zaremba, and Abbeel]{tobin2017domain}
Josh Tobin, Rachel Fong, Alex Ray, Jonas Schneider, Wojciech Zaremba, and Pieter Abbeel.
\newblock Domain randomization for transferring deep neural networks from simulation to the real world.
\newblock In \emph{2017 IEEE/RSJ international conference on intelligent robots and systems (IROS)}, pages 23--30. IEEE, 2017.

\bibitem[Tseng et~al.(2023)Tseng, Li, Kim, Alsisan, Huang, and Kopf]{poseguideddiffusion}
Hung-Yu Tseng, Qinbo Li, Changil Kim, Suhib Alsisan, Jia-Bin Huang, and Johannes Kopf.
\newblock Consistent view synthesis with pose-guided diffusion models.
\newblock In \emph{CVPR}, 2023.

\bibitem[Unterthiner et~al.(2018)Unterthiner, van Steenkiste, Kurach, Marinier, Michalski, and Gelly]{Thomas2018fvd}
Thomas Unterthiner, Sjoerd van Steenkiste, Karol Kurach, Raphael Marinier, Marcin Michalski, and Sylvain Gelly.
\newblock Towards accurate generative models of video: A new metric challenges.
\newblock \emph{arXiv:1812.01717}, 2018.

\bibitem[Wang et~al.(2023{\natexlab{a}})Wang, Yuan, Chen, Zhang, Wang, and Zhang]{wang2023modelscope}
Jiuniu Wang, Hangjie Yuan, Dayou Chen, Yingya Zhang, Xiang Wang, and Shiwei Zhang.
\newblock Modelscope text-to-video technical report.
\newblock \emph{arXiv preprint arXiv:2308.06571}, 2023{\natexlab{a}}.

\bibitem[Wang et~al.(2024{\natexlab{a}})Wang, Ma, Feng, Leng, Yin, and Liang]{wang2024qihoo}
Jing Wang, Ao Ma, Jiasong Feng, Dawei Leng, Yuhui Yin, and Xiaodan Liang.
\newblock Qihoo-t2x: An efficient proxy-tokenized diffusion transformer for text-to-any-task.
\newblock \emph{arXiv preprint arXiv:2409.04005}, 2024{\natexlab{a}}.

\bibitem[Wang et~al.(2023{\natexlab{b}})Wang, Liu, Wang, Li, and Zhang]{streampetr}
Shihao Wang, Yingfei Liu, Tiancai Wang, Ying Li, and Xiangyu Zhang.
\newblock Exploring object-centric temporal modeling for efficient multi-view 3d object detection.
\newblock In \emph{CVPR}, pages 3621--3631, 2023{\natexlab{b}}.

\bibitem[Wang et~al.(2023{\natexlab{c}})Wang, Yang, Tuo, He, Zhu, Fu, and Liu]{wang2023videofactory}
Wenjing Wang, Huan Yang, Zixi Tuo, Huiguo He, Junchen Zhu, Jianlong Fu, and Jiaying Liu.
\newblock Videofactory: Swap attention in spatiotemporal diffusions for text-to-video generation, 2023{\natexlab{c}}.

\bibitem[Wang et~al.(2023{\natexlab{d}})Wang, Zhu, Huang, Chen, Zhu, and Lu]{wang2023drive}
Xiaofeng Wang, Zheng Zhu, Guan Huang, Xinze Chen, Jiagang Zhu, and Jiwen Lu.
\newblock Drivedreamer: Towards real-world-driven world models for autonomous driving.
\newblock \emph{arXiv preprint arXiv:2309.09777}, 2023{\natexlab{d}}.

\bibitem[Wang et~al.(2024{\natexlab{b}})Wang, Chen, Ma, Zhou, Huang, Wang, Yang, He, Yu, Yang, et~al.]{wang2024lavie}
Yaohui Wang, Xinyuan Chen, Xin Ma, Shangchen Zhou, Ziqi Huang, Yi Wang, Ceyuan Yang, Yinan He, Jiashuo Yu, Peiqing Yang, et~al.
\newblock Lavie: High-quality video generation with cascaded latent diffusion models.
\newblock \emph{International Journal of Computer Vision}, pages 1--20, 2024{\natexlab{b}}.

\bibitem[Wen et~al.(2024)Wen, Zhao, Liu, Jia, Wang, Luo, Zhang, Wang, Sun, and Zhang]{wen2024panacea}
Yuqing Wen, Yucheng Zhao, Yingfei Liu, Fan Jia, Yanhui Wang, Chong Luo, Chi Zhang, Tiancai Wang, Xiaoyan Sun, and Xiangyu Zhang.
\newblock Panacea: Panoramic and controllable video generation for autonomous driving.
\newblock In \emph{Proceedings of the IEEE/CVF Conference on Computer Vision and Pattern Recognition}, pages 6902--6912, 2024.

\bibitem[Wu et~al.(2024)Wu, Guo, Tang, Huang, Wang, Chen, and Ding]{drivescape}
Wei Wu, Xi Guo, Weixuan Tang, Tingxuan Huang, Chiyu Wang, Dongyue Chen, and Chenjing Ding.
\newblock Drivescape: Towards high-resolution controllable multi-view driving video generation.
\newblock \emph{arXiv preprint arXiv:2409.05463}, 2024.

\bibitem[Xi et~al.(2025)Xi, Yang, Zhao, Xu, Li, Li, Lin, Cai, Zhang, Li, et~al.]{xi2025sparse}
Haocheng Xi, Shuo Yang, Yilong Zhao, Chenfeng Xu, Muyang Li, Xiuyu Li, Yujun Lin, Han Cai, Jintao Zhang, Dacheng Li, et~al.
\newblock Sparse videogen: Accelerating video diffusion transformers with spatial-temporal sparsity.
\newblock \emph{arXiv preprint arXiv:2502.01776}, 2025.

\bibitem[Yang et~al.(2023)Yang, Ma, Peng, Guo, Lin, and Yu]{bevcontrol}
Kairui Yang, Enhui Ma, Jibin Peng, Qing Guo, Di Lin, and Kaicheng Yu.
\newblock Bevcontrol: Accurately controlling street-view elements with multi-perspective consistency via bev sketch layout.
\newblock \emph{arXiv preprint arXiv:2308.01661}, 2023.

\bibitem[Yang et~al.(2024)Yang, Teng, Zheng, Ding, Huang, Xu, Yang, Hong, Zhang, Feng, et~al.]{cogvideox}
Zhuoyi Yang, Jiayan Teng, Wendi Zheng, Ming Ding, Shiyu Huang, Jiazheng Xu, Yuanming Yang, Wenyi Hong, Xiaohan Zhang, Guanyu Feng, et~al.
\newblock Cogvideox: Text-to-video diffusion models with an expert transformer.
\newblock \emph{arXiv preprint arXiv:2408.06072}, 2024.

\bibitem[Zhang et~al.(2024)Zhang, Li, Le, Shou, Xiong, and Sahoo]{zhang2024moonshot}
David~Junhao Zhang, Dongxu Li, Hung Le, Mike~Zheng Shou, Caiming Xiong, and Doyen Sahoo.
\newblock Moonshot: Towards controllable video generation and editing with multimodal conditions, 2024.

\bibitem[Zhang et~al.(2023)Zhang, Rao, and Agrawala]{zhang2023adding}
Lvmin Zhang, Anyi Rao, and Maneesh Agrawala.
\newblock Adding conditional control to text-to-image diffusion models.
\newblock In \emph{ICCV}, 2023.

\bibitem[Zhang et~al.(2018)Zhang, Ouyang, Li, and Xu]{zhang2018collaborative}
Weichen Zhang, Wanli Ouyang, Wen Li, and Dong Xu.
\newblock Collaborative and adversarial network for unsupervised domain adaptation.
\newblock In \emph{Proceedings of the IEEE conference on computer vision and pattern recognition}, pages 3801--3809, 2018.

\bibitem[Zhao et~al.(2024)Zhao, Wang, Zhu, Chen, Huang, Bao, and Wang]{drivedreamer2}
Guosheng Zhao, Xiaofeng Wang, Zheng Zhu, Xinze Chen, Guan Huang, Xiaoyi Bao, and Xingang Wang.
\newblock Drivedreamer-2: Llm-enhanced world models for diverse driving video generation.
\newblock \emph{arXiv preprint arXiv:2403.06845}, 2024.

\bibitem[Zheng et~al.(2024{\natexlab{a}})Zheng, Song, Guo, Zhang, and Chen]{zheng2024genad}
Wenzhao Zheng, Ruiqi Song, Xianda Guo, Chenming Zhang, and Long Chen.
\newblock Genad: Generative end-to-end autonomous driving.
\newblock \emph{arXiv preprint arXiv:2402.11502}, 2024{\natexlab{a}}.

\bibitem[Zheng et~al.(2024{\natexlab{b}})Zheng, Peng, Yang, Shen, Li, Liu, Zhou, Li, and You]{opensora}
Zangwei Zheng, Xiangyu Peng, Tianji Yang, Chenhui Shen, Shenggui Li, Hongxin Liu, Yukun Zhou, Tianyi Li, and Yang You.
\newblock Open-sora: Democratizing efficient video production for all, 2024{\natexlab{b}}.

\bibitem[Zheng et~al.(2024{\natexlab{c}})Zheng, Peng, Yang, Shen, Li, Liu, Zhou, Li, and You]{zheng2024open}
Zangwei Zheng, Xiangyu Peng, Tianji Yang, Chenhui Shen, Shenggui Li, Hongxin Liu, Yukun Zhou, Tianyi Li, and Yang You.
\newblock Open-sora: Democratizing efficient video production for all.
\newblock \emph{arXiv preprint arXiv:2412.20404}, 2024{\natexlab{c}}.

\bibitem[Zhou et~al.(2023)Zhou, Wang, Yan, Lv, Zhu, and Feng]{zhou2023magicvideo}
Daquan Zhou, Weimin Wang, Hanshu Yan, Weiwei Lv, Yizhe Zhu, and Jiashi Feng.
\newblock Magicvideo: Efficient video generation with latent diffusion models, 2023.

\bibitem[Zhou et~al.(2024)Zhou, Wang, Cai, and Yang]{zhou2024allegro}
Yuan Zhou, Qiuyue Wang, Yuxuan Cai, and Huan Yang.
\newblock Allegro: Open the black box of commercial-level video generation model.
\newblock \emph{arXiv preprint arXiv:2410.15458}, 2024.

\end{thebibliography}
}

\end{document}